\documentclass[sigconf, nonacm]{acmart}

\usepackage{tabularx}
\usepackage{algpseudocode}
\usepackage{tikz}
\usepackage{pgfplots}
\pgfplotsset{compat=1.18}
\usepackage{caption}

\AtBeginDocument{%
  \providecommand\BibTeX{{%
    \normalfont B\kern-0.5em{\scshape i\kern-0.25em b}\kern-0.8em\TeX}}}

\begin{document}

\title{Evaluating the performance-deviation of itemKNN in RecBole and LensKit}

\author{Michael Schmidt}
\email{michael3.schmidt@student.uni-siegen.de}
\affiliation{%
  \institution{University of Siegen}
  \streetaddress{Adolf-Reichwein-Straße 2a}
  \city{Siegen}
  \country{Germany}
  \postcode{57076}
}
\author{Tim Prinz}
\email{tim.prinz@student.uni-siegen.de}
\affiliation{%
  \institution{University of Siegen}
  \streetaddress{Adolf-Reichwein-Straße 2a}
  \city{Siegen}
  \country{Germany}
  \postcode{57076}
}
\author{Jannik Nitschke}
\email{jannik.nitschke@student.uni-siegen.de}
\affiliation{%
  \institution{University of Siegen}
  \streetaddress{Adolf-Reichwein-Straße 2a}
  \city{Siegen}
  \country{Germany}
  \postcode{57076}
}

\renewcommand{\shortauthors}{Schmidt, Prinz, Nitschke}


\begin{abstract}
This study evaluates the performance variations of item-based k-Nearest Neighbors (ItemKNN) algorithms implemented in the recommender system libraries, RecBole and LensKit. By using four datasets (Anime, Modcloth, ML-100K, and ML-1M), we explore the efficiency, accuracy, and scalability of each library's implementation of ItemKNN. The study involves replicating and reproducing experiments to ensure the reliability of results. We are using key metrics such as normalized discounted cumulative gain (nDCG), precision, and recall to evaluate performance with our main focus on nDCG. Our initial findings indicate that RecBole is more performant than LensKit on two out of three metrics. It achieved a
18\% higher nDCG, a 14\% higher Precision and a 35\% lower Recall.
\\
To ensure a fair comparison, we adjusted LensKit's nDCG calculation implementation to match RecBole's approach. After aligning the nDCG calculations implementation, the performance of the two libraries became more comparable. Using implicit feedback, LensKit achieved an nDCG value of 0.2540, whereas RecBole attained a value of 0.2674. Further analysis revealed that the deviations were caused by differences in the implementation of the similarity matrix calculation. Our findings show that RecBole’s implementation outperforms the LensKit algorithm on three out of our four datasets. Following the implementation of a similarity matrix calculation, where only the top K similar items for each item are retained (a method already incorporated in RecBole's ItemKNN), we observed nearly identical nDCG values across all four of our datasets. For example, Lenskit achieved an nDCG value of 0.2586 for the ML-1M dataset with a random seed set to 42. Similarly, RecBole attained the same nDCG value of 0.2586 under identical conditions. Using the original implementation of LensKit's ItemKNN, a higher nDCG value was obtained only on the ModCloth data set.
\end{abstract}

\maketitle

\section{Introduction}
In the age of online shopping and streaming, one thing has become indispensable: recommender systems. Good recommendations lead to satisfied users, making recommender systems central to these services. In this context, Top-N recommendations are frequently mentioned. These are realized by recommender system developers who train models to predict user-specific Top-N recommendations. A commonly used algorithm in such models is the nearest-neighbor algorithm. This algorithm relies on the principle of similarity where each item or user is interpreted as a vector based on its attributes. This allows the calculation (often using the cosine distance) of how similar items or users are, with a small difference implying high similarity.
\\ \\
With the increasing need for effective recommender system, there are now many code libraries available for implementing such algorithms e.g. RecPack \cite{noauthor_recpack_nodate}, Elliot \cite{anelli_elliot_2021}, MyMediaLite \cite{gantner_mymedialite_2011}, LibRec \cite{gupta_auto-caserec_2020} or Auto-CaseRec \cite{guo_librec_nodate}. Two additional libraries are RecBole and LensKit. While LensKit has long been established as a solid machine learning library, RecBole is relatively new. This makes it interesting to determine which library has more efficient algorithms. For our research we used the LensKit 0.14.4 \cite{ekstrand_lenskit_2020} and RecBole 1.2.0 \cite{recbole} versions.
\\ \\
This study aims to examine how the ItemKNN algorithm implemented in LensKit differs from that in RecBole. The commonly used nDCG evaluation metric (normalized discounted cumulative gain) will be used to assess the performance of the algorithms. This metric is particularly suitable for evaluating the quality of Top-N recommendations as it takes into account the position of items in the recommendation list, giving higher scores to relevant items appearing at higher positions. Building on this, we will try to identify differences in their implementation, which, on the one hand, could help practitioners choosing the right library for their purpose, and, on the other hand help developers implement an effective ItemKNN-algorithm.
\\ \\
We begin by simply running both algorithms on one dataset in order to evaluate whether they differ or not and how far. After that we will try to equalize all factors that could lead to different results (e.g. different nDCG calculation in the two libraries). Relying on this, we will use the implementations to explain the different nDCG-values.
\section{Library introduction}
We are working with two libraries, LensKit and RecBole. These are two powerful libraries that are widely used in the field of recommender systems, each offering distinct features to enhance recommendation tasks. 
\\
Starting with LensKit \cite{ekstrand_lenskit_2020}, it was developed by GroupLens researchers at the University of Minnesota. The library comes along with a big and modular framework for constructing, evaluating, and analyzing recommender algorithms. It includes a wide range of recommendation techniques, encompassing collaborative filtering, content-based methods, and hybrid approaches. One of them is the implementation of the k-nearest neighbors (KNN), with parameters such as neighborhood size, similarity metrics, and weighting schemes. LensKit originally had a Java version \cite{ekstrand_rethinking_2011}, but today it is available in Python. There are also extensions of LensKit, such as "LensKit-Auto" \cite{vente_introducing_2023}, which automates the process of selecting and tuning recommendation algorithms for optimal performance.
\\
On the other hand we have RecBole \cite{noauthor_recbole_website}, a library developed by a team of researchers at the Renmin University of China. The library is based on Python and PyTorch in order to reproduce and develop recommendation algorithms in a unified, comprehensive, and efficient framework for research purpose. Although it primarily focuses on deep learning-based recommendation models, it also offers a wide variety of traditional algorithms, like KNN methods. In the KNN module, RecBole integrates advanced features such as neural network-based similarity functions and adaptive neighbor selection strategies. 
\\
Both libraries provide a detailed documentation, tutorials, and examples to show the utilization of their KNN implementations and other functionalities within Python-based recommender systems. 
\section{Method}

\subsection{Data Sets}
We utilized the following four data sets for our experiment: {\textbf{Anime}}, {\textbf{Modcloth}}, {\textbf{ML-100K}} and {\textbf{ML-1M}}. RecBole requires {\textit{"Atomic Files"}} \cite{recbole_atomic_files} to implement their ItemKNN algorithm, which are provided in a Google Drive folder \cite{recbole_google_drive} containing 28 preprocessed datasets ready for direct use by the algorithm. Conversely, LensKit's ItemKNN algorithm can accept any data contained within a pandas DataFrame \cite{lenskit_itemknn_website}. All four data sets were sourced from the Google Drive folder provided by RecBole \cite{recbole_google_drive}.

\subsection{Algorithms}
Since this paper is about the performance-deviation of Item-based k-Nearest Neighbors, we used the implementations of ItemKNN from the LensKit and RecBole libraries for our experiments to evaluate their performance differences \cite{noauthor_recbole_website, ekstrand_lenskit_2020}. Our goal was to highlight the difference between the two Recommender System Libraries, rather than optimizing for the nDCG score, we set the hyperparameter k to a standard value of 20 for simplicity. Each algorithm was then configured to generate 10 recommendations for each user in the test data set. 

\subsection{Pre-processing and Data Splitting}
\label{data-splitting}
In terms of pre-processing, we converted the four data sets into implicit feedback. For the {\textit{ML-100K}}, {\textit{ML-1M}} and {\textit{Modcloth}} data sets we converted ratings higher than three to 1 and lower to 0 and only kept the instances which got a value of 1 (table \ref{table:dataset_statistics_new}). 
This approach is eligible by the following reason: if an item is consistently rated negatively by all users who rated the item, it is unlikely to be recommended to users in the test set. This is because an item deemed irrelevant by the majority of users is not expected to hold significance for any specific user. Another notable consideration is that the corpus of items available for recommendation to a user expands with the inclusion of items that may not align with the user's interests, as evidenced by low ratings from other users.
\newline
Since {\textit{Anime}} has a range of -1 to 10 for their rating values (-1 is used if the user has seen the anime but did not leave a rating for it) \cite{anime_dataset_information} we converted rated items with a rating higher or equal than 6 to 1 and lower than 6 to 0. The reason we did that is the same as above for {\textit{ML-100K}}, {\textit{ML-1M}} and {\textit{Modcloth}}.
\newline
\newline
In order to make it easier to use the splitted data of RecBole for LensKit, we used a 80/20 holdout split with a user-based splitting. User-based splitting is used to handle data sparsity and provides a good amount of data to create recommendations \cite{kant_merging_2018}. In chapter \ref{First_Steps} we used a random seed set to 42. In our later experiments, we used three random seeds, 21, 42 and 84, in order to make sure, that the results not only depend on the random seed \cite{wegmeth_effect_nodate}.
After applying these configurations in RecBole implementation we saved the output in separate folders based on the data sets, so both algorithms (ItemKNN RecBole \& LensKit) get the same data input for their training and the creation of recommendations in the test set.
\newline
\newline
In chapter \ref{Further_Investigations} of our study, we employed the same splitting technique as utilized in the replication phase, with the addition of two distinct random seeds, specifically set to 21 and 84. This approach was adopted to ensure that our experimental results are representative and reproducible.

\begin{table}[h!]
\centering
\captionsetup{justification=raggedright, singlelinecheck=false, labelfont=bf, textfont=normalfont}
\caption{Data set statistics before pre-processing}
\label{table:dataset_statistics_old}

\small
\begin{tabularx}{8.5cm}{|l|X|X|X|X|}
  \hline
   & Anime & Modcloth & ML-100K & ML-1M \\
  \hline
  Users & 73.516 & 47.959 & 944 & 6.041 \\
  Items & 11.201 & 1.379 & 1.683 & 3.707 \\
  Interactions & 7.813.737 & 82.790 & 100.000 & 1.000.209 \\
  Avg. Interactions per user & 106,29 & 1,73 & 106,04 & 165,6 \\
  Avg. Interaction per item & 697,66 & 60,08 & 59,45 & 269,89 \\
  Sparsity & 99.05\% & 99.87\% & 93.71\% & 95.53\% \\
  \hline
\end{tabularx}
\end{table}

\begin{table}[h!]
\centering
\captionsetup{justification=raggedright, singlelinecheck=false, labelfont=bf, textfont=normalfont}
\caption{Data set statistics after pre-processing}
\caption*{Details regarding the data sets post-conversion for implicit feedback}
\label{table:dataset_statistics_new}

\small
\begin{tabularx}{8.5cm}{|l|X|X|X|X|}
  \hline
   & Anime & Modcloth & ML-100K & ML-1M \\
  \hline
  Users & 69.494 & 36.779 & 942 & 6.038 \\
  Items & 9.210 & 1.274 & 1.447 & 3.533 \\
  Interactions & 5.868.892 & 56.722 & 55.375 & 575.281 \\
  Avg. Interactions per user & 84,45 & 1,542 & 58,78 & 95,28 \\
  Avg. Interaction per item & 637,23 & 44,56 & 38,27 & 162,83 \\
  Sparsity & 99.08\% & 99.88\% & 95.94\% & 97.30\% \\
  \hline
\end{tabularx}
\end{table}

\subsection{Algorithm Training and Evaluation}
After splitting the data (details see section \ref{data-splitting}), the train set for each dataset was used to train the RecBole and LensKit ItemKNN algorithms to generate the desired models. The evaluation was realized by using the test sets to create predictions for the users within those and used the nDCG@10 to evaluate the predictions.

\subsection{Hardware Specifications}
The experiments were conducted on a personal computer with the following specifications:
\begin{itemize}
    \item \textbf{Operating System:} Microsoft Windows 11 Home, Version 10.0.22631 Build 22631
    \item \textbf{CPU:} AMD Ryzen 7 5800X, 8 Cores and 16 Threads
    \item \textbf{RAM:} 32GB DDR4
    \item \textbf{GPU:} Nvidia GeForce RTX 3070, 8 GB GDDR6
\end{itemize}
These specifications provided a good amount of computational power to handle the data pre-processing, model training, and evaluation tasks efficiently.
The generation of recommendations for the {\textit{ML-100K}}, {\textit{ML-1M}} and {\textit{Modcloth}} data sets was completed in less than one minute for both libraries, even after adjustment of the ItemKNN of LensKit. For the {\textit{Anime}} data set, RecBole required only one minute to generate recommendations, whereas LensKit took four minutes.
\section{Results}

\subsection{First Steps}
\label{First_Steps}
In this chapter, we tried to get a first impression of the differences in the itemKNN algorithms of RecBole and LensKit. For this purpose, we utilized the well-known ml-100k data set to run both algorithms\cite{repo}. RecBole provides a quickstart routine that allows training, running, and evaluating a model with just one line of code, which significantly simplifies the process. LensKit, on the other hand, required more lines of code and a more complex setup.

\begin{table}[h!]
\centering
\captionsetup{justification=raggedright, singlelinecheck=false, labelfont=bf, textfont=normalfont}
\caption{First results of the two itemKNN algorithms (Random Seed set to 42)}
\label{table:first_statistics}

\small
\begin{tabularx}{8.5cm}{|l|X|X|X|}
  \hline
   & nCDG@10 & Precision@10 & Recall@10\\
  \hline
  LensKit & 0.3067 & 0.2749 & 0.3206 \\
  RecBole & 0.3747 & 0.3187 & 0.2086 \\
  Performance Difference & 18\% & 14\% & 35\% \\
  \hline
\end{tabularx}
\end{table}
\noindent
It is easy to see, that RecBole outperformed LensKit in two out of three metrics: it achieved a 18\% higher nDCG, a 14\% higher Precision but a 35\% lower Recall. These results indicated that there had to be a difference between the two algorithms (or at least the error metrics calculations), motivating us to investigate further.

\subsection{Further Investigations}
\label{Further_Investigations}
\subsubsection{Adjustment of LensKit nDCG Calculation}

After observing the discrepancy in the nDCG values between RecBole and LensKit, we decided to look further into the specific implementation details of each library. Our initial step was to align the nDCG calculations to ensure consistency across both algorithms. The nDCG metric is important in order to evaluate the ranking quality of a recommender system, and even slight differences in the implementation can lead to varying results.
\newpage
\noindent RecBole’s nDCG calculation follows the standard formula{\cite{discounted_2024}}:

\[
nDCG@k = \frac{DCG@k}{IDCG@k}
\]

where

\[
DCG@k = \sum_{i=1}^{k} \frac{2^{rel_i} - 1}{\log_2(i + 1)}
\]

and

\[
IDCG@k = \sum_{i=1}^{|REL_k|} \frac{2^{rel_i} - 1}{\log_2(i + 1)}
\]

\noindent In contrast, the LensKit implementation differs slightly, resulting in different nDCG values compared to RecBole: The difference is that if fewer than k items have been rated, RecBole calculates the IDCG using the last valid value, which would be IDCG@5 for 5 items. In contrast, LensKit still uses IDCG@k. In order to ensure a fair head to head comparison, we modified LensKit's nDCG calculation to match RecBole’s approach.\cite{noauthor_rucaiboxrecbole_2024}\cite[Reproduction folder]{repo}
\\
\noindent After making these adjustments, we ran the evaluations on all our data sets.
Table \ref{table:lenskit_results} shows, that the nDCG value of LensKit for the {\textit{ML-100K}} data set changed, but is still 5\% lower than RecBole's (Figure \ref{fig:ndcg_ml100k}, Random Seed 42). This result indicates, that it was important to isolate the causes of the differences observed in the initial results and provided the baseline for our further analysis of the algorithmic implementations. The resulting adjustments in the LensKit ItemKNN implementation, which are described in the following sections, and the use of the RecBole implementation lead to a foundation of consistent evaluation metrics. All of our implementations can be found in our public GitHub repository \cite{repo}.

\begin{table}[h!]
\centering
\captionsetup{
    justification=raggedright,
    singlelinecheck=false,
    labelfont=bf,
    textfont=normalfont
}
\caption{LensKit nDCG@10 results}
\label{table:lenskit_results}

\small
\begin{tabularx}{\linewidth}{|l|X|X|X|X|}
  \hline
  \textbf{} & \textbf{21} & \textbf{42} & \textbf{84} & Avg.\\
  \hline
  \textbf{ML-100K} & 0.2477 & 0.2540 & 0.2484 & 0.2500\\
  \hline
  \textbf{ML-1M} & 0.2261 & 0.2261 & 0.2274 & 0.2263\\
  \hline
  \textbf{Anime} & 0.2971 & 0.2989 & 0.2970 & 0.2977\\
  \hline
  \textbf{Modcloth} & 0.0992 & 0.0978 & 0.1001 & 0.0990\\
  \hline
\end{tabularx}
\end{table}
\noindent
The RecBole results can be found in figures \ref{fig:ndcg_ml100k}-\ref{fig:ndcg_modcloth}.

\subsubsection{Implementation Difference}
\label{implementation_difference}
To understand the (still) different nDCG values of the two libraries, we examined their implementations. We went through the entire process of data splitting, training, and evaluation and found the following difference in the similarity calculation:
\\ \\
Both libraries use the cosine distance to calculate the similarity of two items. However, while LensKit includes the similarity values to all other items for each item in the similarity matrix \cite{lenskit}, RecBole directly limits the entries per item to the predefined relevant number of neighbors (k) \cite{recboleGit}. This means that a column in the similarity matrix in RecBole contains the k nearest neighbors for an item with their similarity values. All other entries in the column are set to zero, resulting in \textit{item-count minus k} zero-entries.\\ \\
For example, when we have 1000 items, both algorithms start by constructing a 1000x1000 similarity matrix. Now, both use the cosine distance to calculate the distance from item 1 to all other items contained in the similarity matrix. LensKit now takes all of the calculated values to fill the first column in the matrix. RecBole on the other hand, filters for the topK values in the 1000 similarities and puts just these topK values in the column. All other entries are set to zero. This is done for every column.
\\ \\
This difference becomes apparent in the prediction of items for a user. The prediction algorithm in both libraries is essentially the same: based on the user-specific "rated"-corpus (the interactions of the user in the test set), the relevance is calculated for each item known by the algorithm, meaning all items from the train corpus. LensKit does it as follows \cite{lkpyItem}:
\\ 
\begin{algorithmic}
\For{$item$ \textbf{in} $train\_unique\_items$}
\State $sims = [\ ]$
    \For{$ratedItem$ in $ratedByUser$}
        \State{$sims.append(similarityMatrix[item][ratedItem])$}
    \EndFor
    \State{$topKSims = topK(sims)$}
    \State{$itemScore = sum(sims)$}
\EndFor
\end{algorithmic}\leavevmode\\
\noindent
Now, a score is assigned to each item, and the items with the highest scores can be recommended. RecBole essentially does the same using matrix multiplication. But since there are many zero-entries in RecBole's similarity matrix, the array \textit{sims} will contain many zeros. Only these matrix-entries, where \textit{ratedItem} is in the topK similarities of the \textit{item}, will have a score higher than zero.

\subsubsection{Possible advantages of the RecBole-Implementation}
Besides saving memory (the matrix is stored in CSR format), RecBole's implementation may have a major advantage: reducing the number of column entries to k similarity values per item minimizes noise during score calculation in the prediction process. In LensKit, even dissimilar items can be weighted if they are in the topK of all rated items. This cannot happen in RecBole: the item must be among the k nearest neighbors, otherwise it is weighted with zero.

\subsubsection{Adjustment of LensKit ItemKNN}
\begin{figure}[h!]
    \centering
    \captionsetup{justification=raggedright, singlelinecheck=false, labelfont=bf, textfont=normalfont}
    \caption{Comparison of nDCG@10 for ML-100K dataset}
    \includegraphics[]{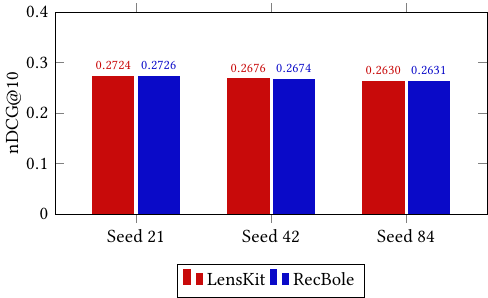}
    \label{fig:ndcg_ml100k}
\end{figure}

\begin{figure}[h!]
    \centering
    \captionsetup{justification=raggedright, singlelinecheck=false, labelfont=bf, textfont=normalfont}
    \caption{Comparison of nDCG@10 for ML-1M dataset}
    \includegraphics[]{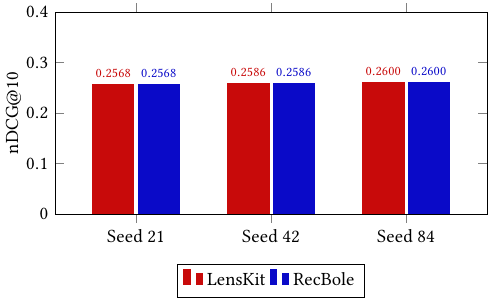}
    \label{fig:ndcg_ml1m}
\end{figure}

\begin{figure}[h!]
    \centering
    \captionsetup{justification=raggedright, singlelinecheck=false, labelfont=bf, textfont=normalfont}
    \caption{Comparison of nDCG@10 for Anime dataset}
    \includegraphics[]{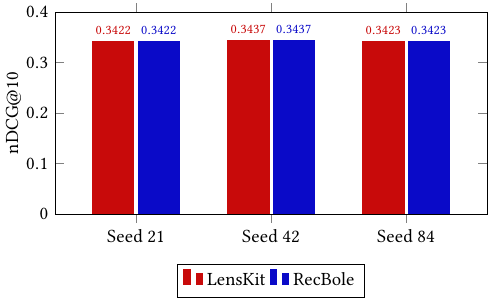}
    \label{fig:ndcg_anime}
\end{figure}

\begin{figure}[h!]
    \centering
    \captionsetup{justification=raggedright, singlelinecheck=false, labelfont=bf, textfont=normalfont}
    \caption{Comparison of nDCG@10 for Modcloth dataset}
    \includegraphics[]{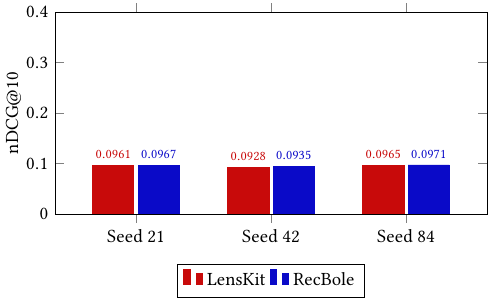}
    \label{fig:ndcg_modcloth}
\end{figure}

To explain the differences in the nDCG results after equalizing the nDCG implementation and providing the LensKit algorithm with the same data as RecBole, we analyzed the ItemKNN implementations of both algorithms. By decomposing and understanding both algorithms, we gained a more precise perspective. In doing so, we identified a significant difference in the creation of the similarity matrix during the training (more details see \ref{implementation_difference}). Recognizing this disparity, we modified LensKit's ItemKNN implementation to generate a similarity matrix akin to that of RecBole. Subsequently, we re-evaluated both algorithms using the previously mentioned four data sets (ML-100K, ML-1M, Anime and Modcloth) and three random seeds (21, 42 and 84). The results presented in figures \ref{fig:ndcg_ml100k}-\ref{fig:ndcg_modcloth} illustrate the outcomes of these adjustments. It is clear to see that after the adjustments, both models almost got the same nDCG@10 results for every data set and random seed provided.

\newpage

\newpage

\subsection{Discussion}
Our findings show that RecBole's implementation outperforms the LensKit algorithm on three out of our four data sets. Only on the ModCloth data set LensKit's original implementation achieves an nDCG of 0.0978 (more details see table \ref{table:lenskit_results}), while the adjusted implementation results in a nDCG of 0.0935 (see figure \ref{fig:ndcg_modcloth}).
\\ \\
Our first adjustment involved standardizing the nDCG calculation across both implementations. This adjustment was crucial for ensuring a fair comparison and revealed that the initial performance deviations were partly due to differences in the nDCG calculation methods. Once aligned, the results showed more comparable performance metrics between the two libraries. This highlights the importance of standardized evaluation metrics.
\\ \\
Relying on the now comparable nDCG calculation, we conducted a deeper investigation and, by deeply studying both implementations, discovered the part causing the different nDCG values: the similarity-matrix calculation. After adjusting the similarity-matrix calculation in LensKit's source code\cite[Code folder]{repo}, the nDCG values became nearly equal to RecBole's and got better in three out of four cases.
\\ \\
Our results indicate, that adopting RecBole's implementation could lead to better overall performance. However, to determine which implementation is superior or more suitable for specific cases, further experiments are necessary. These should include a wider variety of data sets with different characteristics, such as varying levels of sparsity, diversity in item types, and different user interaction patterns.
\\ \\
Still, one can see that we were able to boost LensKit's nDCG values in three out of four cases by adjusting the similarity-matrix calculation. We believe RecBole's approach, which limits the entries per column in the similarity matrix to k items, effectively ignores irrelevant items. This could result in more accurate item-scores, leading to more accurate predictions.
\\ \\
\newline
In conclusion, we demonstrated that different similarity-matrix calculations lead to different predictions, even when using the same prediction algorithm, which can cause significant variations in the nDCG results. While we cannot definitively state that RecBole's implementation is better, an important takeaway is that developers implementing the ItemKNN algorithm should carefully consider the similarity-matrix calculation and its impact on prediction performance.

\section{Acknowledgements}
This work was conducted as part of a Machine Learning Internship  2024 at the University of Siegen, Department of Electrical Engineering and Computer Science, Intelligent Systems Group \cite{Beel2024}.

\bibliographystyle{ACM-Reference-Format}
\bibliography{references}


\begin{thebibliography}{23}

%%% the Web address.

\ifx \showCODEN    \undefined \def \showCODEN     #1{\unskip}     \fi
\ifx \showDOI      \undefined \def \showDOI       #1{#1}\fi
\ifx \showISBNx    \undefined \def \showISBNx     #1{\unskip}     \fi
\ifx \showISBNxiii \undefined \def \showISBNxiii  #1{\unskip}     \fi
\ifx \showISSN     \undefined \def \showISSN      #1{\unskip}     \fi
\ifx \showLCCN     \undefined \def \showLCCN      #1{\unskip}     \fi
\ifx \shownote     \undefined \def \shownote      #1{#1}          \fi
\ifx \showarticletitle \undefined \def \showarticletitle #1{#1}   \fi
\ifx \showURL      \undefined \def \showURL       {\relax}        \fi
% invisible to TeX
\providecommand\bibfield[2]{#2}
\providecommand\bibinfo[2]{#2}
\providecommand\natexlab[1]{#1}
\providecommand\showeprint[2][]{arXiv:#2}

\bibitem[noa(2022)]%
        {noauthor_recpack_nodate}
 \bibinfo{year}{2022}\natexlab{}.
\newblock \bibinfo{title}{{RecPack}: {An}(other) {Experimentation} {Toolkit} for {Top}-{N} {Recommendation} using {Implicit} {Feedback} {Data} {\textbar} {Proceedings} of the 16th {ACM} {Conference} on {Recommender} {Systems}}.
\newblock
\newblock
\urldef\tempurl%
\url{https://dl.acm.org/doi/abs/10.1145/3523227.3551472}
\showURL{%
\tempurl}


\bibitem[dis(2024)]%
        {discounted_2024}
 \bibinfo{year}{2024}\natexlab{}.
\newblock \bibinfo{title}{Discounted cumulative gain}.
\newblock
\newblock
\urldef\tempurl%
\url{https://en.wikipedia.org/w/index.php?title=Discounted_cumulative_gain&oldid=1223546723#Normalized_DCG}
\showURL{%
\tempurl}
\newblock
\shownote{Page Version ID: 1223546723}.


\bibitem[noa(2024)]%
        {noauthor_rucaiboxrecbole_2024}
 \bibinfo{year}{2024}\natexlab{}.
\newblock \bibinfo{title}{{RUCAIBox}/{RecBole}}.
\newblock
\newblock
\urldef\tempurl%
\url{https://github.com/RUCAIBox/RecBole/blob/1.2.x/recbole/evaluator/metrics.py}
\showURL{%
\tempurl}
\newblock
\shownote{original-date: 2020-06-11T15:18:11Z}.


\bibitem[len(6 05)]%
        {lenskit}
 \bibinfo{year}{2024-06-05}\natexlab{}.
\newblock \bibinfo{title}{{LensKit} {ItemKNN} {Model} {Repository} [v0.14.4]}.
\newblock
\newblock
\urldef\tempurl%
\url{https://github.com/lenskit/lkpy/blob/0.14.4/lenskit/algorithms/item_knn.py}
\showURL{%
\tempurl}


\bibitem[rec(6 05)]%
        {recboleGit}
 \bibinfo{year}{2024-06-05}\natexlab{}.
\newblock \bibinfo{title}{{RecBole} {ItemKNN} {Model} {Repository} [v1.2.0]}.
\newblock
\newblock
\urldef\tempurl%
\url{https://github.com/RUCAIBox/RecBole/blob/v1.2.0/recbole/model/general_recommender/itemknn.py}
\showURL{%
\tempurl}


\bibitem[rec(6 10)]%
        {recbole_google_drive}
 \bibinfo{year}{2024-06-10}\natexlab{}.
\newblock \bibinfo{title}{{RecBole} – {Google} {Drive}}.
\newblock
\newblock
\urldef\tempurl%
\url{https://drive.google.com/drive/folders/1so0lckI6N6_niVEYaBu-LIcpOdZf99kj}
\showURL{%
\tempurl}


\bibitem[rec(6 18)]%
        {recbole_atomic_files}
 \bibinfo{year}{2024-06-18}\natexlab{}.
\newblock \bibinfo{title}{Atomic {Files} — {RecBole} 1.2.0 documentation}.
\newblock
\newblock
\urldef\tempurl%
\url{https://recbole.io/docs/user_guide/data/atomic_files.html}
\showURL{%
\tempurl}


\bibitem[len(6 18)]%
        {lenskit_itemknn_website}
 \bibinfo{year}{2024-06-18}\natexlab{}.
\newblock \bibinfo{title}{k-{NN} {Collaborative} {Filtering} — {LensKit} 0.14.4 documentation}.
\newblock
\newblock
\urldef\tempurl%
\url{https://lkpy.readthedocs.io/en/stable/knn.html}
\showURL{%
\tempurl}


\bibitem[lkp(6 18)]%
        {lkpyItem}
 \bibinfo{year}{2024-06-18}\natexlab{}.
\newblock \bibinfo{title}{{LensKit} {ItemKNN} user prediction [v0.14.4]}.
\newblock
\newblock
\urldef\tempurl%
\url{https://github.com/lenskit/lkpy/blob/4dfa41073760af430d074bfb671e482767348c12/lenskit/algorithms/item_knn.py#L447}
\showURL{%
\tempurl}


\bibitem[rep(6 18)]%
        {repo}
 \bibinfo{year}{2024-06-18}\natexlab{}.
\newblock \bibinfo{title}{Our {Repository}: {itemKnn}-{LensKit}-vs-{Recbole}}.
\newblock
\newblock
\urldef\tempurl%
\url{https://github.com/GoogleMichMal/itemKnn-LensKit-vs-Recbole/tree/main}
\showURL{%
\tempurl}


\bibitem[noa(6 18)]%
        {noauthor_recbole_website}
 \bibinfo{year}{2024-06-18}\natexlab{}.
\newblock \bibinfo{title}{{RecBole}}.
\newblock
\newblock
\urldef\tempurl%
\url{https://recbole.io/}
\showURL{%
\tempurl}


\bibitem[Anelli et~al\mbox{.}(2021)]%
        {anelli_elliot_2021}
\bibfield{author}{\bibinfo{person}{Vito~Walter Anelli}, \bibinfo{person}{Alejandro Bellogín}, \bibinfo{person}{Antonio Ferrara}, \bibinfo{person}{Daniele Malitesta}, \bibinfo{person}{Felice~Antonio Merra}, \bibinfo{person}{Claudio Pomo}, \bibinfo{person}{Francesco~Maria Donini}, {and} \bibinfo{person}{Tommaso Di~Noia}.} \bibinfo{year}{2021}\natexlab{}.
\newblock \showarticletitle{Elliot: a {Comprehensive} and {Rigorous} {Framework} for {Reproducible} {Recommender} {Systems} {Evaluation}}. In \bibinfo{booktitle}{\emph{Proceedings of the 44th {International} {ACM} {SIGIR} {Conference} on {Research} and {Development} in {Information} {Retrieval}}}. \bibinfo{pages}{2405--2414}.
\newblock
\urldef\tempurl%
\url{https://doi.org/10.1145/3404835.3463245}
\showDOI{\tempurl}
\newblock
\shownote{arXiv:2103.02590 [cs]}.


\bibitem[Beel et~al\mbox{.}(2024)]%
        {Beel2024}
\bibfield{author}{\bibinfo{person}{Joeran Beel}, \bibinfo{person}{Tobias Vente}, {and} \bibinfo{person}{Lukas Wegmeth}.} \bibinfo{year}{2024}\natexlab{}.
\newblock \showarticletitle{Praktikum Maschinelles Lernen}.
\newblock \bibinfo{journal}{\emph{Universisät Siegen}} (\bibinfo{year}{2024}).
\newblock


\bibitem[Drive(6 18)]%
        {anime_dataset_information}
\bibfield{author}{\bibinfo{person}{RecBole~Google Drive}.} \bibinfo{year}{2024-06-18}\natexlab{}.
\newblock \bibinfo{title}{anime-example – {Google} {Drive}}.
\newblock
\newblock
\urldef\tempurl%
\url{https://drive.google.com/drive/folders/1bBAbfK0znXpN28xg0Et1Vn9Ln7UTzxCG}
\showURL{%
\tempurl}


\bibitem[Ekstrand(2020)]%
        {ekstrand_lenskit_2020}
\bibfield{author}{\bibinfo{person}{Michael~D. Ekstrand}.} \bibinfo{year}{2020}\natexlab{}.
\newblock \showarticletitle{{LensKit} for {Python}: {Next}-{Generation} {Software} for {Recommender} {Systems} {Experiments}}. In \bibinfo{booktitle}{\emph{Proceedings of the 29th {ACM} {International} {Conference} on {Information} \& {Knowledge} {Management}}}. \bibinfo{publisher}{ACM}, \bibinfo{address}{Virtual Event Ireland}, \bibinfo{pages}{2999--3006}.
\newblock
\showISBNx{978-1-4503-6859-9}
\urldef\tempurl%
\url{https://doi.org/10.1145/3340531.3412778}
\showDOI{\tempurl}


\bibitem[Ekstrand et~al\mbox{.}(2011)]%
        {ekstrand_rethinking_2011}
\bibfield{author}{\bibinfo{person}{Michael~D. Ekstrand}, \bibinfo{person}{Michael Ludwig}, \bibinfo{person}{Joseph~A. Konstan}, {and} \bibinfo{person}{John~T. Riedl}.} \bibinfo{year}{2011}\natexlab{}.
\newblock \showarticletitle{Rethinking the recommender research ecosystem: reproducibility, openness, and {LensKit}}. In \bibinfo{booktitle}{\emph{Proceedings of the fifth {ACM} conference on {Recommender} systems}}. \bibinfo{publisher}{ACM}, \bibinfo{address}{Chicago Illinois USA}, \bibinfo{pages}{133--140}.
\newblock
\showISBNx{978-1-4503-0683-6}
\urldef\tempurl%
\url{https://doi.org/10.1145/2043932.2043958}
\showDOI{\tempurl}


\bibitem[Gantner et~al\mbox{.}(2011)]%
        {gantner_mymedialite_2011}
\bibfield{author}{\bibinfo{person}{Zeno Gantner}, \bibinfo{person}{Steffen Rendle}, \bibinfo{person}{Christoph Freudenthaler}, {and} \bibinfo{person}{Lars Schmidt-Thieme}.} \bibinfo{year}{2011}\natexlab{}.
\newblock \showarticletitle{{MyMediaLite}: a free recommender system library}. In \bibinfo{booktitle}{\emph{Proceedings of the fifth {ACM} conference on {Recommender} systems}} \emph{(\bibinfo{series}{{RecSys} '11})}. \bibinfo{publisher}{Association for Computing Machinery}, \bibinfo{address}{New York, NY, USA}, \bibinfo{pages}{305--308}.
\newblock
\showISBNx{978-1-4503-0683-6}
\urldef\tempurl%
\url{https://doi.org/10.1145/2043932.2043989}
\showDOI{\tempurl}


\bibitem[Guo et~al\mbox{.}({[n.\,d.]})]%
        {guo_librec_nodate}
\bibfield{author}{\bibinfo{person}{Guibing Guo}, \bibinfo{person}{Jie Zhang}, \bibinfo{person}{Zhu Sun}, {and} \bibinfo{person}{Neil Yorke-Smith}.} \bibinfo{year}{[n.\,d.]}\natexlab{}.
\newblock \showarticletitle{{LibRec}: {A} {Java} {Library} for {Recommender} {Systems}}.
\newblock  (\bibinfo{year}{[n.\,d.]}).
\newblock
\urldef\tempurl%
\url{https://ceur-ws.org/Vol-1388/demo_paper1.pdf}
\showURL{%
\tempurl}


\bibitem[Gupta and Beel(2020)]%
        {gupta_auto-caserec_2020}
\bibfield{author}{\bibinfo{person}{Srijan Gupta} {and} \bibinfo{person}{Joeran Beel}.} \bibinfo{year}{2020}\natexlab{}.
\newblock \bibinfo{title}{Auto-{CaseRec}: {Automatically} {Selecting} and {Optimizing} {Recommendation}-{Systems} {Algorithms}}.
\newblock
\newblock
\urldef\tempurl%
\url{https://doi.org/10.31219/osf.io/4znmd}
\showDOI{\tempurl}


\bibitem[Kant and Mahara(2018)]%
        {kant_merging_2018}
\bibfield{author}{\bibinfo{person}{Surya Kant} {and} \bibinfo{person}{Tripti Mahara}.} \bibinfo{year}{2018}\natexlab{}.
\newblock \showarticletitle{Merging user and item based collaborative filtering to alleviate data sparsity}.
\newblock \bibinfo{journal}{\emph{International Journal of System Assurance Engineering and Management}} \bibinfo{volume}{9}, \bibinfo{number}{1} (\bibinfo{date}{Feb.} \bibinfo{year}{2018}), \bibinfo{pages}{173--179}.
\newblock
\showISSN{0976-4348}
\urldef\tempurl%
\url{https://doi.org/10.1007/s13198-016-0500-9}
\showDOI{\tempurl}


\bibitem[Vente et~al\mbox{.}(2023)]%
        {vente_introducing_2023}
\bibfield{author}{\bibinfo{person}{Tobias Vente}, \bibinfo{person}{Michael Ekstrand}, {and} \bibinfo{person}{Joeran Beel}.} \bibinfo{year}{2023}\natexlab{}.
\newblock \showarticletitle{Introducing {LensKit}-{Auto}, an {Experimental} {Automated} {Recommender} {System} ({AutoRecSys}) {Toolkit}}. In \bibinfo{booktitle}{\emph{Proceedings of the 17th {ACM} {Conference} on {Recommender} {Systems}}} \emph{(\bibinfo{series}{{RecSys} '23})}. \bibinfo{publisher}{Association for Computing Machinery}, \bibinfo{address}{New York, NY, USA}, \bibinfo{pages}{1212--1216}.
\newblock
\showISBNx{9798400702419}
\urldef\tempurl%
\url{https://doi.org/10.1145/3604915.3610656}
\showDOI{\tempurl}


\bibitem[Wegmeth et~al\mbox{.}(2023)]%
        {wegmeth_effect_nodate}
\bibfield{author}{\bibinfo{person}{Lukas Wegmeth}, \bibinfo{person}{Tobias Vente}, \bibinfo{person}{Lennart Purucker}, {and} \bibinfo{person}{Joeran Beel}.} \bibinfo{year}{2023}\natexlab{}.
\newblock \showarticletitle{The {Effect} of {Random} {Seeds} for {Data} {Splitting} on {Recommendation} {Accuracy}}.
\newblock  (\bibinfo{year}{2023}).
\newblock


\bibitem[Zhao et~al\mbox{.}(2021)]%
        {recbole}
\bibfield{author}{\bibinfo{person}{Wayne~Xin Zhao}, \bibinfo{person}{Shanlei Mu}, \bibinfo{person}{Yupeng Hou}, \bibinfo{person}{Zihan Lin}, \bibinfo{person}{Yushuo Chen}, \bibinfo{person}{Xingyu Pan}, \bibinfo{person}{Kaiyuan Li}, \bibinfo{person}{Yujie Lu}, \bibinfo{person}{Hui Wang}, \bibinfo{person}{Changxin Tian}, \bibinfo{person}{Yingqian Min}, \bibinfo{person}{Zhichao Feng}, \bibinfo{person}{Xinyan Fan}, \bibinfo{person}{Xu Chen}, \bibinfo{person}{Pengfei Wang}, \bibinfo{person}{Wendi Ji}, \bibinfo{person}{Yaliang Li}, \bibinfo{person}{Xiaoling Wang}, {and} \bibinfo{person}{Ji{-}Rong Wen}.} \bibinfo{year}{2021}\natexlab{}.
\newblock \showarticletitle{RecBole: Towards a Unified, Comprehensive and Efficient Framework for Recommendation Algorithms}. In \bibinfo{booktitle}{\emph{{CIKM}}}. \bibinfo{publisher}{{ACM}}, \bibinfo{pages}{4653--4664}.
\newblock


\end{thebibliography}

\end{document}